\documentclass[a4paper, amsfonts, amssymb, amsmath, onecolumn, showkeys, nofootinbib, twoside, superscriptaddress,longbibliography]{revtex4-2}
\usepackage[english]{babel}
\usepackage[utf8]{inputenc}
\usepackage[colorinlistoftodos, color=green!40, prependcaption]{todonotes}
\usepackage{cancel,amsmath}
\usepackage{amsthm}
\usepackage{mathtools}
\usepackage{xcolor}
\usepackage{graphicx}
\usepackage{float}
\usepackage{soul}

\usepackage{gensymb}
\usepackage{multirow}
\usepackage[left=23mm,right=13mm,top=35mm,columnsep=15pt]{geometry} 
\usepackage{adjustbox}
\usepackage{placeins}
\usepackage[T1]{fontenc}

\usepackage{csquotes}

\usepackage[pdftex, pdftitle={Article}, pdfauthor={Author}]{hyperref} % For hyperlinks in the PDF

\usepackage{changes}
\usepackage[capitalize]{cleveref}

\begin{document}
\title{First-Passage Approach to Optimizing Perturbations for Improved Training of Machine Learning Models}

\author{Sagi Meir}
    \affiliation{School of Chemistry, Tel Aviv University, Tel Aviv 6997801, Israel.}
    \affiliation{The Center for Physics and Chemistry of Living Systems, Tel Aviv University, Tel Aviv 6997801, Israel.}
\author{Tommer D. Keidar}
    \affiliation{School of Chemistry, Tel Aviv University, Tel Aviv 6997801, Israel.}
    \affiliation{The Center for Physics and Chemistry of Living Systems, Tel Aviv University, Tel Aviv 6997801, Israel.}
\author{Shlomi Reuveni}
    \affiliation{School of Chemistry, Tel Aviv University, Tel Aviv 6997801, Israel.}
    \affiliation{The Center for Physics and Chemistry of Living Systems, Tel Aviv University, Tel Aviv 6997801, Israel.}
    \affiliation{The Center for Computational Molecular and Materials Science, Tel Aviv University, Tel Aviv 6997801, Israel.}
\author{Barak Hirshberg\footnote{Author to whom any correspondence should be addressed hirshb@tauex.tau.ac.il}}
    % \email{hirshb@tauex.tau.ac.il}
    \affiliation{School of Chemistry, Tel Aviv University, Tel Aviv 6997801, Israel.}
    \affiliation{The Center for Physics and Chemistry of Living Systems, Tel Aviv University, Tel Aviv 6997801, Israel.}
    \affiliation{The Center for Computational Molecular and Materials Science, Tel Aviv University, Tel Aviv 6997801, Israel.}

\begin{abstract}
    Machine learning models have become indispensable tools in applications across the physical sciences. Their training is often time-consuming, vastly exceeding the inference timescales. Several protocols have been developed to perturb the learning process and improve the training, such as shrink and perturb, warm restarts, and stochastic resetting. For classifiers, these perturbations have been shown to result in enhanced speedups or improved generalization. However, the design of such perturbations is usually done \textit{ad hoc} by intuition and trial and error. To rationally optimize training protocols, we frame them as first-passage processes and consider their response to perturbations. 
    We show that if the unperturbed learning process reaches a quasi-steady state, the response at a single perturbation frequency can predict the behavior at a wide range of frequencies. 
    We employ this approach to a CIFAR-10 classifier using the ResNet-18 model and identify a useful perturbation and frequency among several possibilities. 
    {We demonstrate the transferability of the approach to other datasets, architectures, optimizers and even tasks (regression instead of classification).}
    Our work allows optimization of perturbations for improving the training of machine learning models using a first-passage approach.
\end{abstract}


\maketitle

\section{Introduction}
Machine learning algorithms and deep neural network (NN) models have become powerful tools across chemistry and physics to tackle challenges that were once computationally or experimentally prohibitive. Examples include protein structure prediction~\cite{jumper2021highly}, entropy calculation for physical systems~\cite{nir2020machine}, machine learning potentials~\cite{behler2007generalized}, and many others~\cite{carleo2017solving, noe2019boltzmann, mate2024neural, anelli2024robust, lagemann2021deep, ravuri2021skilful, wang2020machine, rupp2012fast, tsai2020learning, geiger2013neural}. Unfortunately, these capabilities come with a time-consuming price spent on model training. Second-order optimization algorithms are unfeasible for deep NN~\cite{fukumizu2000local,loshchilov2016sgdr}, and currently, stochastic gradient descent (SGD){~\cite{rumelhart1986learning, peng2019accelerating, loshchilov2016sgdr}} and its variants{~\cite{duchi2011adaptive,zeiler2012adadelta, kingma2014adam}}, remain a cornerstone of NN training~\cite{nguyen2019machine, sutskever2013importance}. 

SGD by itself can be slow and prone to getting stuck in local minima. Several common practices are used to address these limitations. First is the incorporation of momentum and adaptive learning rates as done by modern optimizers, usually leading to faster convergence and improved performance~\cite{sutskever2013importance, zeiler2012adadelta, kingma2014adam, leimkuhler2019partitioned}. The second, which is the focus of this work, is to ``perturb'' the training process, which may lead to speedup or improved generalization. Examples include shrink \& perturb (S\&P)~\cite{ash2020warm, zaidi2023does}, warm restarts~\cite{loshchilov2016sgdr}, stochastic resetting (SR)~\cite{bae2025stochastic} and continual backpropagation~\cite{dohare2024loss}. 

Ash et al.~\cite{ash2020warm} first suggested the S\&P protocol to improve generalization during online learning. More recently, Zaidi et al.~\cite{zaidi2023does} investigated when such re-initialization might help the training, and concluded that while it is clear S\&P is helpful in some cases, a general theory of why it works is missing. Bae et al.~\cite{bae2025stochastic} applied SR  to a dynamically-updated checkpoint during the training of an NN with a resetting time sampled from an exponential distribution. They concluded that their strategy decreases overfitting on noisy labels and leads to better generalization. Yet, predicting the optimal resetting rate a priori was not possible. Loshchilov et al.~\cite{loshchilov2016sgdr} showed that a protocol with warm restarts converges up to four times faster to the same test accuracy compared to standard SGD training. 
Overall, despite their usefulness, a theory capable of predicting the effect of such perturbation protocols is missing, and their design is often done by intuition and empirical trial and error. 

The resemblance of SGD to Langevin dynamics~\cite{feng2021inverse,cheng2020stochastic, jules2023charting} has been utilized to describe and analyze the learning process of NNs with concepts and methods from thermodynamics{~\cite{alemi2018therml}} and statistical physics~\cite{seung1992statistical, choromanska2015loss, zdeborova2020understanding, karniadakis2021physics, carleo2019machine}.
For example, Feng et al. found an inverse fluctuation-dissipation relation between weights' fluctuations and the flatness of the loss landscape. Based on their finding, they developed an algorithm that delays catastrophic forgetting in sequential learning tasks~\cite{feng2021inverse}.

Another example is the work of Stephan et al. where they showed that SGD with a constant learning rate can be used as an approximate Bayesian posterior inference algorithm~\cite{stephan2017stochastic}. Their result was obtained by viewing SGD as a Markov chain with a stationary distribution.
However, despite the advancements in applying statistical mechanical tools in machine learning, they have not been previously used to design perturbations to improve the performance of SGD optimizers.

Recently, Keidar et al. developed a response theory to predict how rare perturbations affect the completion of an arbitrary stochastic process~\cite{keidar2024universal}. Their theory focused on first-passage processes~\cite{Redner_2001,metzler2014first,bray2013persistence}, in which a stochastic process has a predefined distribution of initial conditions and a well-defined target. The stochastic nature of the process gives rise to a distribution of first-passage times, i.e., the first time the process reaches its target. In this work, we treat the SGD-based training of NNs up to a predefined target accuracy as a first-passage process. We treat protocols such as S\&P, warm restarts, and stochastic resetting as perturbations to the first-passage process that occur every $P$ epochs during the training. We employ the theory of Keidar et al. for the first time to analyze and improve the training of NN models using such perturbations. We show that, given a set of protocols, we can determine which would lead to the highest acceleration and identify the optimal perturbation time interval $P$. We focus on three types of protocols: S\&P, partial re-initialization of small weights (partial SR), and full SR. Finally, we suggest a methodology for testing and analyzing new perturbation protocols.
{We demonstrate the transferability of the approach across datasets (CIFAR-10, CIFAR-100~\cite{krizhevsky2009learning} and MNIST~\cite{deng2012mnist}), architectures (ResNet-18~\cite{he2016deep} and fully-connected NN), optimizers (SGD, SGD with momentum, Adam~\cite{kingma2014adam}) and even tasks (classification and regression) in the supporting information.}

\section{Theory}

\begin{figure}[h]
\begin{center}
\centerline{\includegraphics[width=0.7\linewidth]{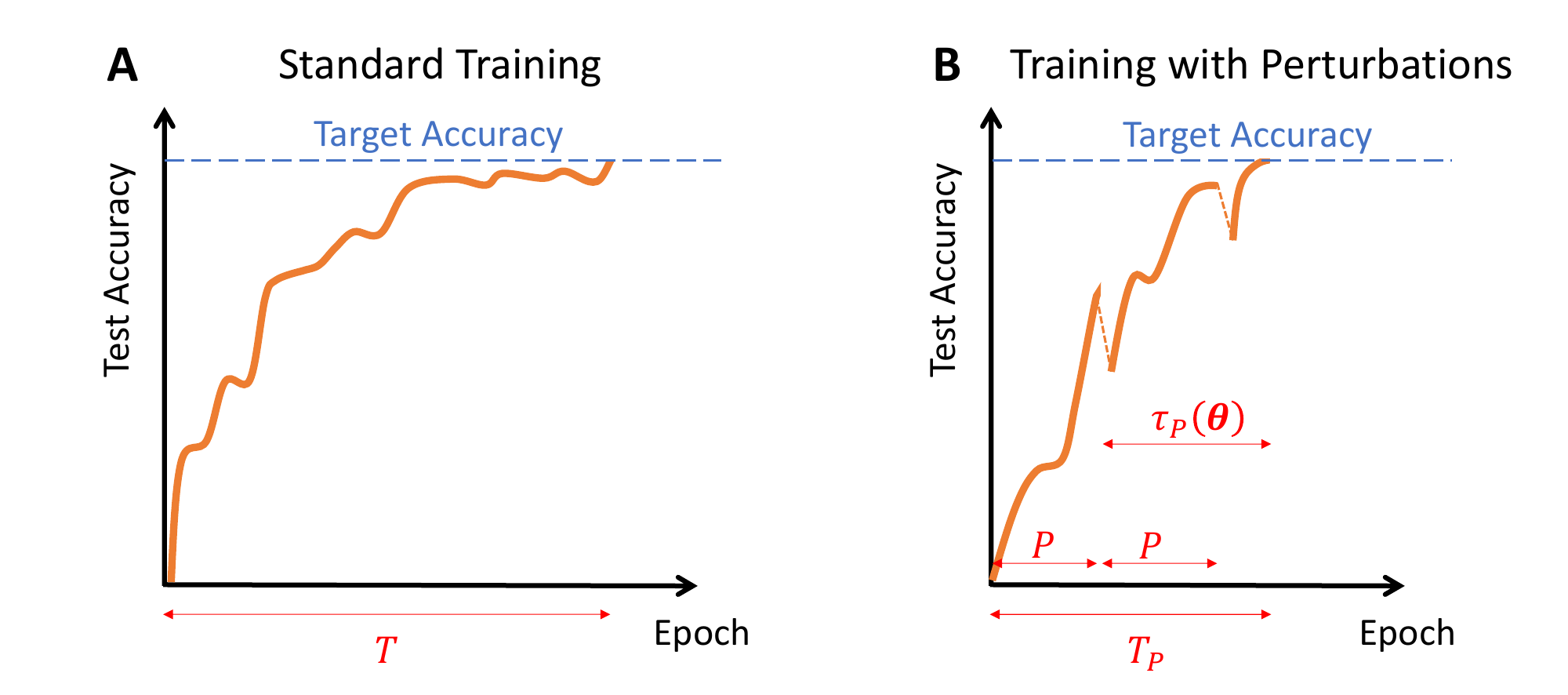}}
\caption{Training NNs as a first-passage process. Panel A presents the test accuracy as a function of the number of epochs for training without perturbations (orange line). Here, $T$ is the FPT to reach the target accuracy (dashed blue line). Panel B presents the test accuracy as a function of the number of epochs for training with a perturbation every $P$ epochs (orange line). $T_P$ is the perturbed FPT to the target accuracy (dashed blue line) and $\tau_P(\boldsymbol{\theta})$ is the residual time to reach the target accuracy after the first perturbation.}
\label{fig:LR_Theory}
\end{center}
\end{figure}

In this work, we treat the NN training in the absence of a perturbation as a stochastic process that is characterized by a propagator $G(\boldsymbol{\theta},t)$, representing the probability of being in state $\boldsymbol{\theta}$ at time $t$. We emphasize that $\boldsymbol{\theta}$ characterizes the overall state of the system, i.e., it may represent the weights, biases, hyperparameters, etc. We focus on first-passage processes (see \cref{fig:LR_Theory}A), and define the first-passage time (FPT)  as the first instance in which the test accuracy reaches a certain threshold, which is treated as an absorbing boundary condition. Namely, once the test accuracy of a model reaches the target, training is stopped. Since it is a stochastic process, there will be an FPT distribution. We denote with $T$ the random variable representing the FPT of the unperturbed training process. We define the survival probability {$\Psi_T(t)\equiv\Pr(T>t)$} as the fraction of models which did not reach the threshold at time $t$. {Note, that $\Psi_T(t)$ is one minus the cumulative distribution function (CDF) of $T$. In terms of the propagator, the survival probability can be written as}
\begin{equation}
    \Psi_T(t) = \int_\Theta G(\boldsymbol{\theta},t)\mathrm{d}\boldsymbol{\theta},
\label{eq:Psi-definition}
\end{equation}
where $\Theta$ symbolizes that the integration domain is over all possible states.

\cref{fig:LR_Theory}B shows the training process with a perturbation that is applied every $P$ epochs. We denote by $T_P$ the random variable representing the FPT of the perturbed process to the target accuracy. The perturbation can be of any form, e.g., it can affect the network weights, hyperparameters, activation functions, or any other component of the training. Keidar et al. showed that the perturbed FPT is connected to the unperturbed FPT through~\cite{keidar2024universal},
\begin{equation}
T_P= \begin{cases}T & \text { if } T \leq P, \\ P+\tau_P(\boldsymbol{\theta}) & \text { if } T>P,\end{cases}
\label{eq:T_P_cases_description}
\end{equation}
where $\tau_P(\boldsymbol{\theta})$ is the random variable representing the residual number of epochs it takes to reach the target after the perturbation has been first applied (see \cref{fig:LR_Theory}B). It depends on $\boldsymbol{\theta}$, the state of the network at time $P$, which might be different for every realization of the process. 

Applying the law of total expectation to \cref{eq:T_P_cases_description} (see the derivation in the supporting information), we obtain the mean FPT under perturbations, $\mathbb{E}[T_P]$,
\begin{equation}
    \mathbb{E}[T_P]=\sum_{t=0}^{P-1}\Psi_T(t)  + \Psi_T(P) \Bar{\tau}_{P},
\label{eq:LR-prediction}
\end{equation}
where $\Bar{\tau}_{P}$ is given by
\begin{equation}
    \Bar{\tau}_{P} = \int_{\Theta} \mathbb{E}[\tau_P(\boldsymbol{\theta})] \frac{G(\boldsymbol{\theta},P)}{\Psi_T(P)} \mathrm{d}\boldsymbol{\theta}.
\label{eq:tau_P-definition}
\end{equation}
In \cref{eq:tau_P-definition}, $\mathbb{E}[\tau_P(\boldsymbol{\theta})]$ is the average residual time after the first perturbation over all possible noise realizations of the stochastic training process after the perturbation was applied to a given state $\boldsymbol{\theta}$ at time $P$.  Hence, $\Bar{\tau}_{P}$ can be understood as the average of $\mathbb{E}[\tau_P(\boldsymbol{\theta})]$ over all possible $\boldsymbol{\theta}$ generated by the unperturbed process at time $P$ right before the perturbation has been applied.

\cref{eq:LR-prediction} formally decomposes the mean FPT with the perturbation to two contributions. The first term, $\sum_{t=0}^{P-1}\Psi_T(t)$, sets a lower bound on $\mathbb{E}[T_P]$ that only depends on the unperturbed process. The second term encodes all the effects of the perturbation on the mean FPT. These equations allow, in principle to predict the effect of a perturbation on the training for all $P$. 

Consider, for example, the specific case of SR as the perturbation{~\cite{reuveni2016optimal, pal2017first, evans2020stochastic, blumer2022stochastic, blumer2024combining}}. Then, every $P$ epochs, the state of the system is restarted by resampling the initial conditions for training. As a result, for any training process, $\tau_P(\boldsymbol{\theta})$ does not depend on the state at time $P$, and is simply an independent and identically distributed copy of $T_P$. In that case, $\Bar{\tau}_P=\mathbb{E}[T_P]$, and substituting it into \cref{eq:LR-prediction} gives~\cite{eliazar2020mean},
\begin{equation}
    \mathbb{E}[T_P]_{SR}=\frac{1}{1-\Psi_T(P)}\sum_{t=0}^{P-1}\Psi_T(t) .
\label{eq:SR-prediction}
\end{equation}
\cref{eq:SR-prediction} shows that, for SR, the mean FPT with the perturbation can be predicted entirely from the survival probability of the unperturbed learning process. 
To derive \cref{eq:SR-prediction}, we used the specific properties of SR. For other perturbations, one must assume something about the dynamics of the underlying training process as we do in the next section.

\section{Results and Discussion}

\subsection{Quasi-steady state leads to $P$-independent residual times}
\label{sec:qss-theory}
\begin{figure}[h]
\begin{center}
\centerline{\includegraphics[width=0.69\linewidth]{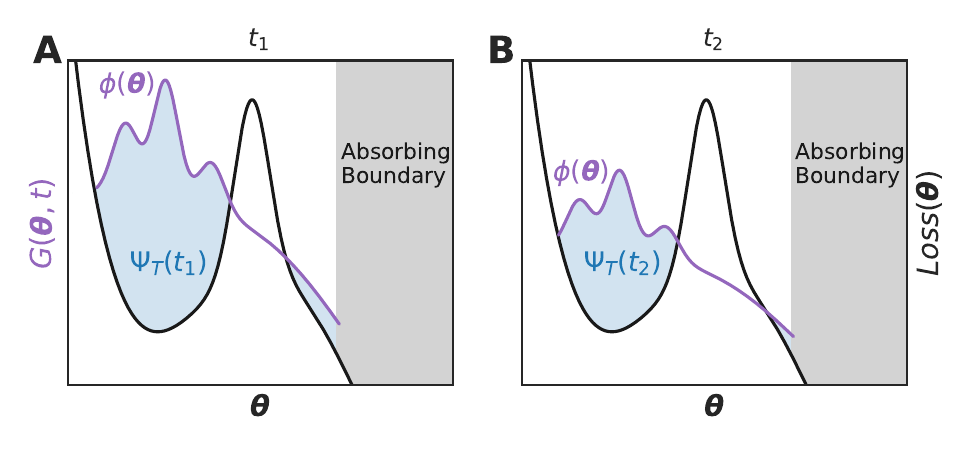}}
\caption{An illustration of a QSS of a propagator $G(\boldsymbol{\theta},t)$ at times $t_2>t_1>t_r$. At QSS, the probability density function $\phi(\boldsymbol{\theta})$ (purple lines) maintains the same shape, while the fraction of models that survived (blue areas) declines over time through panels A-B, i.e., $\Psi_T(t_1)>\Psi_T(t_2)$.}
\label{fig:qss_drawing}
\end{center}
\end{figure}

To proceed beyond \cref{eq:LR-prediction}, we consider a stochastic process in which the propagator reaches a quasi-steady state (QSS) after a typical relaxation time $t_r$,
\begin{equation}
    G(\boldsymbol{\theta},t)\simeq \phi(\boldsymbol{\theta})\Psi_T(t) \quad \text { for } t>t_r.
\label{eq:quasi-steady state}
\end{equation}
In \cref{eq:quasi-steady state}, $\phi(\boldsymbol{\theta})$ is a time-independent probability density function of the network state $\boldsymbol{\theta}$. In other words, a QSS is defined by constant relative populations of all states $\boldsymbol{\theta}$, although the overall number of models that survived declines over time through $\Psi_T(t)$~\cite{nitzan2024chemical}, as illustrated in \cref{fig:qss_drawing}. The QSS approximation is commonly applied to analyze chemical kinetics~\cite{nitzan2024chemical, ji2021stiff, kramers1940brownian, hanggi1986escape, hanggi1990reaction}, but has not been used to explain NN training before. 

{A QSS could also emerge during the training of certain NNs, for example, through a separation of timescales in learning. Typically, the training loss drops rapidly in the early stages of learning, and a much slower decrease in loss is observed at later stages while training accuracy plateaus~\cite{feng2021inverse}. 
In other words, the state of the network evolves rapidly at first, but later the model approaches a local minimum of the loss landscape leading to smaller gradient updates of the parameters. This suggests that beyond a characteristic relaxation time, the state of the network becomes approximately stationary. 
Next, we show that in this case, i.e., when a QSS emerges, the mean residual time $\Bar{\tau}_P$ becomes independent of the perturbation epoch $P$, provided it is rare. As a result, we can predict the mean first-passage time (FPT) across a wide range of perturbation frequencies by sampling the training process at a single frequency.}

% {Our motivation that a QSS emerges in the training process of some NNs is driven by the empirical time scale separation of learning. This separation is manifested by an initial rapid decrease in the training loss and an increase in the training accuracy, followed by a slower decrease of the training loss while the training accuracy is roughly steady at 100\%~\cite{feng2021inverse}. In other words, during the first few epochs, the state of the network tremendously varies, in contrast to the rest of the training where the model approaches a local minimum leading to smaller gradient updates of the parameters. It means that after a typical time scale, the state of the model is nearly constant.} 

To show this, we consider three timescales in the system: the relaxation time $t_r$, the perturbation time interval $P$ and the mean residual time $\Bar{\tau}_P$. Considering a perturbation that is done at a time $P > t_r$, and utilizing QSS, we observe that the perturbation acts on a state sampled from $\phi(\boldsymbol{\theta})$ irrespective of $P$. We can then plug \cref{eq:quasi-steady state} into \cref{eq:tau_P-definition} and get
\begin{equation}
    \Bar{\tau}_{P} = \int_{\Theta} \mathbb{E}[\tau_P(\boldsymbol{\theta})] \phi(\boldsymbol{\theta}) \mathrm{d}\boldsymbol{\theta}.
\label{eq:tau-indep-P_step1}
\end{equation}
Note that the mean residual time for completion after the perturbation is applied still depends on $P$ through $\mathbb{E}[\tau_P(\boldsymbol{\theta})]$. Next, if the perturbation is rare enough, i.e., $P > \Bar{\tau}_P$, we effectively perturb the system only once on average before the training is complete. As a result, $\tau_P(\boldsymbol{\theta})=\tau(\boldsymbol{\theta})$, i.e., does not depend on $P$, leading to
\begin{equation}
    \Bar{\tau}_{P} \approx \Bar{\tau} = \int_{\Theta} \mathbb{E}[\tau(\boldsymbol{\theta})] \phi(\boldsymbol{\theta}) \mathrm{d}\boldsymbol{\theta} \quad \text { for } P>\max\left(t_r, \Bar{\tau}_P\right).
\label{eq:tau-indep-P}
\end{equation}
%\textcolor{red}{Note that we obtained \cref{eq:tau-indep-P} by assuming a QSS, but hypothetically there could be non-QSS scenarios for which \cref{eq:tau-indep-P} would hold. Fortunately, our method applies also to such cases.} 
Plugging \cref{eq:tau-indep-P} into \cref{eq:LR-prediction} leads to
\begin{equation}
    \mathbb{E}[T_P]=\sum_{t=0}^{P-1}\Psi_T(t)  + \Psi_T(P) \Bar{\tau}.
\label{eq:LR-prediction-rarepert}
\end{equation}
\cref{eq:LR-prediction-rarepert} sets the recipe for designing useful perturbations and the frequencies at which to apply them. It tells us to perform simulations applying the perturbation only once at some $P^*>\max\left(t_r, \Bar{\tau}_P\right)$. These simulations provide the unbiased survival function until time $P^*$ and $\Bar{\tau}$. Combining the two, we can predict the mean FPT for all values $\max\left(t_r, \Bar{\tau}_P\right) \leq P \leq P^*$. 

Below, we first test the QSS hypothesis when training a CIFAR-10~\cite{krizhevsky2009learning} classifier using the ResNet-18 model~\cite{he2016deep}. Then, for two perturbations, S\&P and partial SR, we show that indeed $\Bar{\tau}_{P} \approx \Bar{\tau}$ for a wide range of $P$. Finally, we use \cref{eq:LR-prediction-rarepert} to predict the mean FPT under these perturbations at a wide range of $P$, and benchmark them against brute-force training. We show that this procedure can be used to select a perturbation and time interval $P$ that lead to faster training.

\subsection{Experimental test of the quasi-steady-state}

A naive approach to demonstrate a QSS when training an NN model would be to sample several learning trajectories while keeping track of the parameters to obtain $G(\boldsymbol{\theta},t)$. However, it is impractical to keep track of millions of parameters, and the finite sample size introduces noise to the estimation of the distribution of $G(\boldsymbol{\theta},t)$. This noise increases with the dimensionality of $\boldsymbol{\theta}$, making it infeasible to demonstrate a QSS convergence from the trajectory data of $\boldsymbol{\theta}$. 
Instead, we draw inspiration from enhanced sampling of free energy surfaces~\cite{barducci2011metadynamics, valsson2014variational, bussi2020using} and define a collective variable,  $A(\boldsymbol{\theta})$, representing the state of the system and look at its marginal distribution, 
\begin{equation}
    G(A,t) = \int_{\Theta} G(\boldsymbol{\theta},t)\delta(A(\boldsymbol{\theta})-A)\mathrm{d}\boldsymbol{\theta}.
\label{eq:acc_propagator-def}
\end{equation}
In practice, we will use the test set accuracy as a collective variable. 
If $G(\boldsymbol{\theta},t)$ reaches a QSS, and we plug \cref{eq:quasi-steady state} into \cref{eq:acc_propagator-def}, we get a QSS also in $G(A,t)$,
\begin{equation}
    G(A,t) \simeq \Psi_T(t)\int_{\Theta} \phi(\boldsymbol{\theta})\delta(A(\boldsymbol{\theta})-A)\mathrm{d}\boldsymbol{\theta} = \Psi_T(t)\phi(A), \quad \text { for } t>t_r.
\label{eq:acc_quasi-steay}
\end{equation}
In \cref{eq:acc_quasi-steay}, $\phi(A)=\int_{\Theta} \phi(\boldsymbol{\theta})\delta(A(\boldsymbol{\theta})-A)\mathrm{d}\boldsymbol{\theta}$ is the time-independent probability density function of $A$. Therefore, we will use the dynamics of $G(A,t)$ as a proxy to justify the QSS hypothesis.

\begin{figure}[h]
\centering
    \begin{minipage}[t]{0.78\linewidth}
    \centering
    \includegraphics[width=\linewidth]{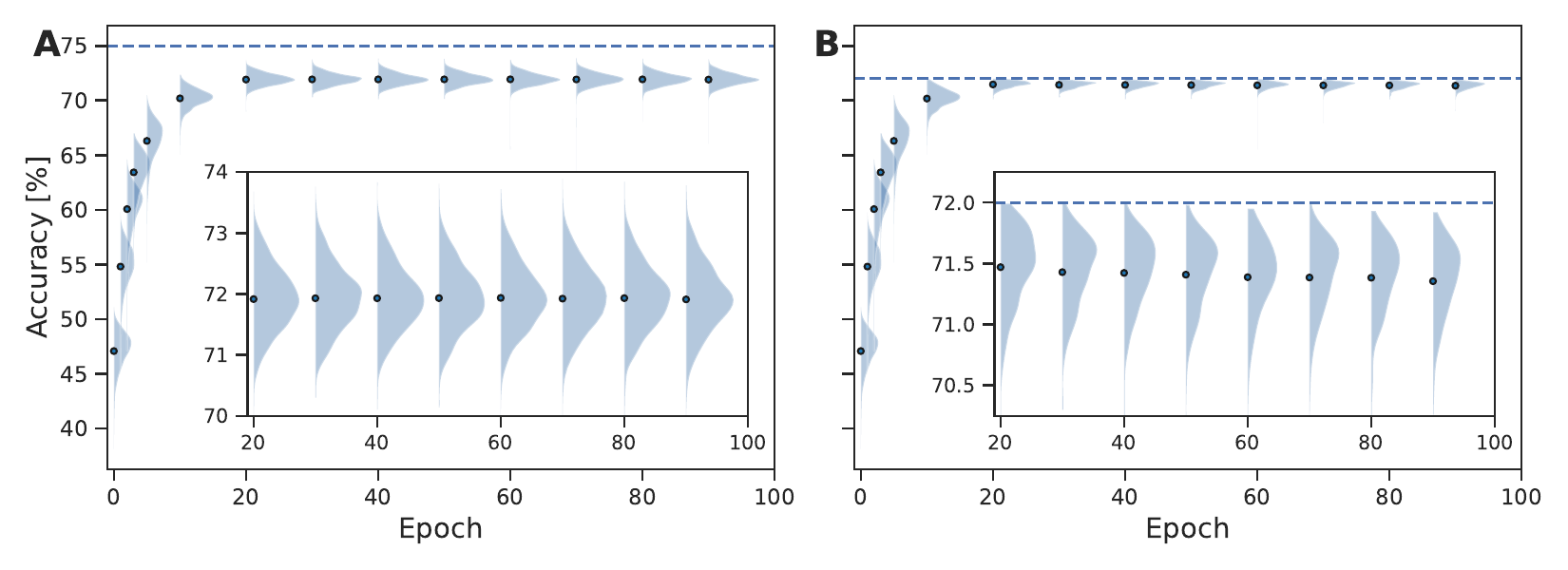}
    \end{minipage}
    \hspace{-0.0262\linewidth}
    \begin{minipage}[t]{0.2195\linewidth}
    \centering
    \includegraphics[width=\linewidth]{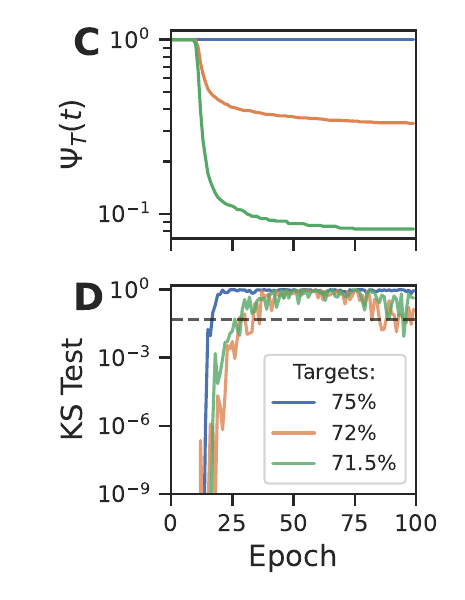}
    \end{minipage}
\caption{Experimental evidence for a quasi-steady-state. Panels A and B are violin plots of the density distributions of the test accuracy as a function of the number of epochs, for models that did not reach the target. Blue dots represent the mean accuracy and the blue areas are the distributions rotated by $90^{\circ}$. The dashed blue lines are the target accuracies, set to 75\% and 72\% in panels A and B, respectively. In both cases, after $\sim$20-30 epochs, the system reaches a QSS, i.e., $G(A,t)\simeq \phi(A)\Psi_T(t)$. 
Panel A is a special case in which $\Psi_T(t)=1$, while in panel B, $\Psi_T(t)$ decays slowly. 
Panel C shows the survivals $\Psi_T(t)$ for the 75\% (blue), 72\% (orange), and 71.5\% (green) target test accuracies. 
Panel D is the Kolmogorov–Smirnov (KS) test between the CDFs of the accuracy $A$ at different epochs to the average CDF of $A$ over epochs 20 to 100. 
}
\label{fig:acc_densities_over_time}
\end{figure}

We used the ResNet-18 architecture as our NN model for the classification task of the CIFAR-10 dataset. We trained 1000 models using random initialization and plotted the test accuracy as a function of the number of epochs, which we will refer to as trajectories. We used standard SGD training, without perturbations. See the Computational Details section for the full training setup. To obtain the propagator $G(A,t)$ from the trajectory data for a specific target test accuracy, we treat it as an absorbing boundary. In other words, trajectories that reached the target accuracy are stopped and excluded from $G(A,t)$ at later times. The survival $\Psi_T(t)$ is estimated by the fraction of models that did not reach the target at epoch $t$ of the training, {i.e.,
\begin{equation}\label{eq:_survival_est}
    \Psi_T(t) = \Pr(T>t) \approx \frac{1}{N}\sum_{i=1}^{N} \mathcal{I}_i(t),
\end{equation}
where $N$ is the total number of sampled trajectories and $\mathcal{I}_i(t)$ is an indicator function which gives 1 if the $i$-th trajectory did not reach the target before time $t$.}
In \cref{fig:acc_densities_over_time}A, we start by analyzing the specific case of a target test accuracy of 75\% (dashed blue line). For this accuracy, no model reaches the target up to 100 epochs, i.e., $\Psi_T(t)=1 \, \forall t$ (blue line \cref{fig:acc_densities_over_time}C). The violins of \cref{fig:acc_densities_over_time}A represent the density distribution of $A$ as a function of time. It narrows down with time, until epoch $\sim20$, and then remains with a fixed shape and height, which suggests that $G(A,t)=\phi(A)$ from that epoch onwards. 
%We define the relaxation time according to the convergence of the mean, i.e., $t_r\sim20$ epochs. 
Besides this qualitative shape comparison, we also quantitatively compare the cumulative distribution functions (CDFs) of $A$ to the average CDF of $A$ over epochs 20 to 100 with the Kolmogorov–Smirnov (KS) test~\cite{massey1951kolmogorov, tiwary2015role, blumer2024short}, shown in \cref{fig:acc_densities_over_time}D (blue line). We define the relaxation time as having a p-value larger than 0.05 in the KS test, which gives a relaxation time of $t_r=17$ epochs. Alternative statistical tests result in similar relaxation times (see the supporting information).

In \cref{fig:acc_densities_over_time}B we set the target test accuracy to 72\% which leads to a decay of $\Psi_T(t)$ with the number of epochs, i.e., some trajectories reach the target during the training (orange line in \cref{fig:acc_densities_over_time}C). Although there is a decline in the number of models that did not reach the target, after $\sim30$ epochs the shape of the density distribution again remains approximately fixed while the height decreases, as expected from a QSS (\cref{eq:acc_quasi-steay}). 
%Additional, indication to a QSS is the orange line in \cref{fig:acc_densities_over_time}C that changes its decay behavior at $t_r$ to a slow decay. 
We compare the CDFs of $A$ to the average CDF of $A$ over epochs 20 to 100 with the KS test~\cite{massey1951kolmogorov, tiwary2015role, blumer2024short}, conditioned on model survival.
Similarly to the previous case, we plot the p-value of the KS-test in \cref{fig:acc_densities_over_time}D (orange line) and obtain $t_r = 28$ epochs.
We also checked the QSS hypothesis for a lower target accuracy of $71.5\%$ and plot the survival function and KS-test p-value in \cref{fig:acc_densities_over_time}C,D (green lines), respectively. Again, we obtain a similar relaxation time of $t_r = 26$ epochs.

All these examples are consistent with our hypothesis that the training of a CIFAR-10 classifier using a ResNet-18 model reaches a QSS for various target accuracies. {Moreover, we find that the QSS is not a unique feature of the CIFAR-10 dataset and ResNet-18 architecture. We tested a different dataset, network architecture, and optimizer and found a QSS in all cases (See section E in the SI). We even demonstrate a QSS for a regression task similar to that of Dekel et al.~\cite{dekel2022pr}.}

\subsection{Experimental test that the residual time is $P$-independent}
We now show that if $P$ is large enough, introducing a perturbation would result in $P$-independent residual times, according to \cref{eq:tau-indep-P}. Here, we show it experimentally by focusing on the 72\% target test accuracy  case (\cref{fig:acc_densities_over_time}B) where the system reaches a QSS after $t_r=28$ epochs. We analyze the dependence of the residual time to reach the target on the perturbation time $P$. To do so, we take only the models that did not reach the target after $P=100,\,50,\,20,$ and 10 epochs, introduce the perturbation, and continue the training until reaching the target. We consider two perturbations, S\&P and partial SR. 

The S\&P perturbation takes the weights and biases of the NN at the moment it is applied, shrinks them by a factor $\lambda$ and adds an independent and identically distributed copy sampled from the initial conditions distribution, multiplied by a factor $\gamma$. We used the same values as in ref. \cite{zaidi2023does}, $\lambda=0.4$ and $\gamma=0.1$. The partial SR perturbation takes only a fraction of the parameters of the NN and re-initializes them to an independent and identically distributed copy sampled from the initial conditions distribution. {Since small weights may contribute less to the performance of the model, their re-initialization holds the potential of escaping local minima while not throwing away most of the learned weights.} We chose to initialize 30\% of the weights, those with the smallest absolute value at the moment the perturbation is applied. 

\begin{figure}[h]
\begin{center}
\centerline{\includegraphics[width=0.99\linewidth]{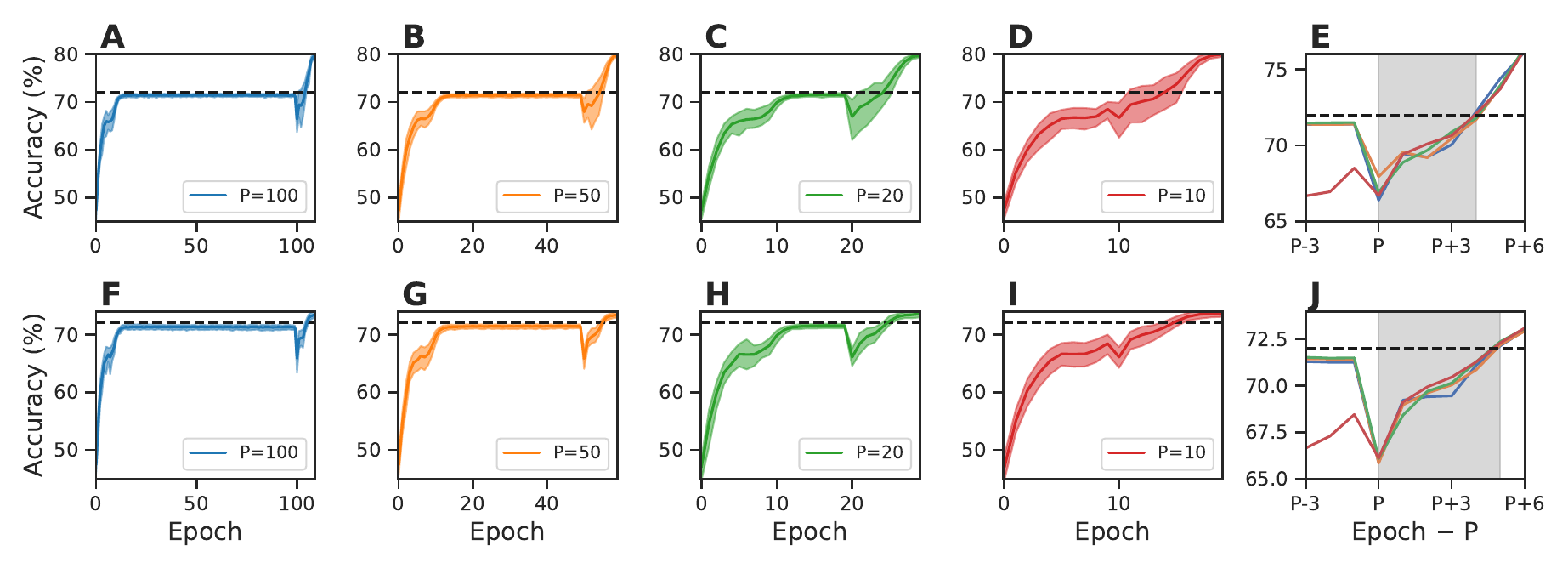}}
\caption{The mean trajectory of the test accuracy and its top and bottom deciles (shaded areas), for models that did not reach the target after $P$ epochs. From left to right $P=100$, $50$, $20$, $10$ (see legend). The dashed black lines represent 72\% test accuracy (the target). At epoch number $P$ we apply the S\&P protocol (upper row, panels A-E) or the partial SR protocol (bottom row, panels F-J). Panels E and J center the mean trajectories with respect to their $P$. The gray areas are from the moment of perturbation until the first time that the average trajectory reaches the target accuracy.}
\label{fig:s_and_p_constant_tau}
\end{center}
\end{figure}

In \cref{fig:s_and_p_constant_tau}, we first plot the mean test accuracy of the models as a function of the epoch. The top and bottom rows correspond to the S\&P and the partial resetting perturbations, respectively. In all cases, we find that the residual time of the mean trajectory to reach the target is smaller than $P$ and that the perturbation is only applied once before reaching the target. 
As a result, when $P>t_r$, we expect that the mean residual time will be $P$-independent, as explained in \cref{sec:qss-theory}. 
 In panels E and J we center all the mean trajectories with respect to their perturbation time $P$. We find that all curves approximately collapse on top of each other regardless of $P$, reaching the target accuracy after the same number of epochs since the perturbation has been applied (gray area). We emphasize that the residual time is different for the two perturbations, as shown in panels E and J, but is $P$-independent in both cases.

\begin{figure}[h]
\begin{center}
\centerline{\includegraphics[width=0.4\linewidth]{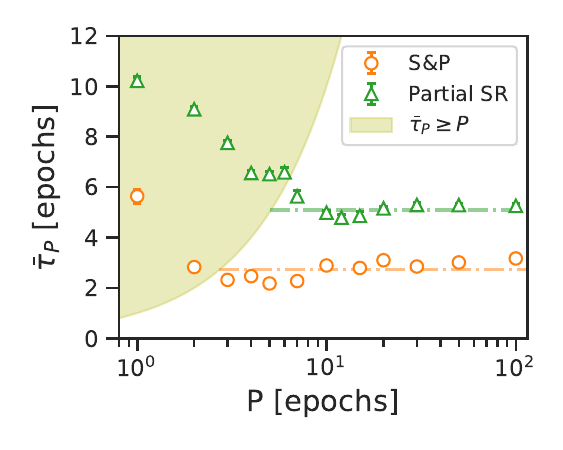}}
\caption{The mean residual time $\Bar{\tau}_P$ versus the perturbation time. The values of $\Bar{\tau}_P$ for S\&P (orange circles) and partial SR (green triangles) are approximately constant when $P>\Bar{\tau}_P$ (outside the yellow area). Dashed dotted lines are averages of all $\Bar{\tau}_P<P$.}
\label{fig:tau_P_vs_P}
\end{center}
\end{figure}

Next, instead of plotting the average training curve, we plot the value of the {mean} residual time $\Bar{\tau}_P$ as a function of $P$ in \cref{fig:tau_P_vs_P} for a larger range of $P=1\text{--}100$. 
We observe that $\Bar{\tau}_P$ is roughly constant over nearly two orders of magnitude in $P$ for both perturbations. Surprisingly, $\Bar{\tau}_P \approx \Bar{\tau}$ even for values of $P$ that are smaller than the relaxation time, beyond the expected range according to \cref{eq:tau-indep-P}. However, when $P$ becomes smaller than the residual time ($\sim3$ and $\sim5$ for S\&P and partial SR, respectively), denoted by the shaded area in \cref{fig:tau_P_vs_P}, the process is perturbed more than once on average before reaching the target and the residual time is no longer $P$-independent. {We observed the same behavior for different datasets and tasks presented in Section E of the supporting information.}

\subsection{Prediction of the mean FPT}
\begin{figure}[h]
\begin{center}
\centerline{\includegraphics[width=0.4\linewidth]{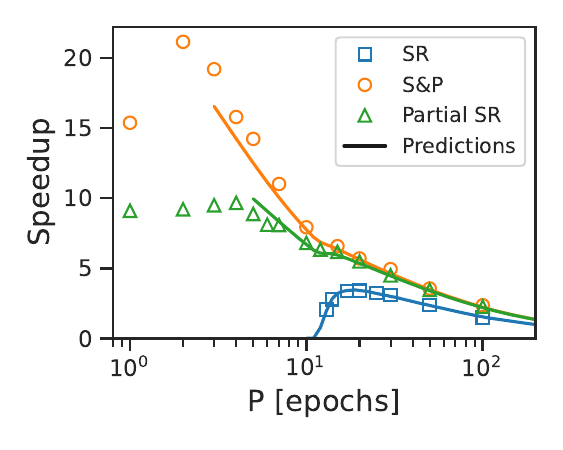}}
\caption{The speedup gained by using different perturbation protocols versus the perturbation time. Orange and green solid lines are the theoretical predictions of \cref{eq:LR-prediction-rarepert} for S\&P and partial SR, respectively. For the predictions, we used the value of $\Bar{\tau}_{P^*}$ for $P^*=100$. We plot the predictions only down to $P=\Bar{\tau}_{P^*}$.
The blue solid line is the theoretical prediction for the SR case (\cref{eq:SR-prediction}). Symbols represent {averages over} brute force training at every value of $P$.}
\label{fig:Speedup}
\end{center}
\end{figure}

In this section, we show how to predict the mean FPT with a perturbation for a wide range of $P$ from as few experiments as possible. 
We begin by assuming that several training trajectories are already available, and we wish to assess whether applying a perturbation is worthwhile, and select at what time interval to apply it. The simplest perturbation to analyze is SR. It comes at no additional cost, through \cref{eq:SR-prediction}, since we already sampled the survival function. To quantify the acceleration in reaching a target accuracy by applying a perturbation, we define the speedup as the ratio between the mean FPT without and with the perturbation, respectively. We plot the predicted speedup for SR in \cref{fig:Speedup} (blue line) and compare it with numerical experiments (blue squares). We find that SR leads to a maximal speedup of $\sim3$ using a perturbation time interval of $\sim20$. 

To go beyond SR to other perturbations that could potentially lead to higher speedups, we must measure their residual time.
A naive approach of simply evaluating $\Bar{\tau}_P$ for every $P$ and using \cref{eq:LR-prediction} would mean running the training for all perturbations at all $P$ values, which is costly. Instead, we would like to use the fact that the residual time is $P$-independent. To that end, we sample the training with the perturbation applied once at some $P^*$ that is large enough. We then use \cref{eq:LR-prediction-rarepert} to predict the mean FPT for a wide range of $P < P^*$. 
We plot the predicted speedup in \cref{fig:Speedup} for the S\&P (orange line) and partial SR (green line) perturbations, using their value of $\Bar{\tau}_{P^*}$ at $P^*=100$. Going from right to left on the $P$ axis, the predictions fit the experimental values of the S\&P (orange circles) and partial SR (green triangles) protocols for $P\geq\Bar{\tau}_{P^*}$, as expected. %{Note that the experimental speedup values are costly because they require averaging over several trained models with and without a perturbation.}

Using these predictions, we identify that S\&P leads to a speedup of $\sim16$ at a perturbation time interval of 3, while partial SR leads to a speedup of $\sim8$ at $P=7$. These predictions correctly identify S\&P as the preferred strategy that leads to the highest speedup. Although the experimental maximal speedup is slightly larger ($\sim21$), our predictions are a computationally efficient way of determining useful perturbations and time intervals that lead to high speedups.

We suggest a general methodology that can examine different kinds of perturbations and pick the optimal one and an efficient $P$. Consider for example a classification task for which the architecture of the NN and the hyperparameters of the SGD-based optimization are given. Then, it is common practice to train in parallel an ensemble of models and terminate any training process that did not reach a minimal desired test accuracy after a tolerance time of $P^*$ epochs. 
After this process is completed, we propose to estimate the survival function, $\Psi_T(t)$ for $t\leq P^*$ {obtained by \cref{eq:_survival_est}}. This is already enough information to evaluate whether SR leads to any acceleration. 
For other perturbations, evaluate first if the training accuracy of the unperturbed process reaches a QSS and determine $t_r < P^*$ using the KS test. If that is the case, introduce several candidate perturbations to the models that did not reach the target accuracy and measure $\Bar{\tau}_{P^*}$ for each one. 
This way, $\Psi_T(t\leq P^*)$ and $\Bar{\tau}_{P^*}$ are obtained simultaneously for several perturbation protocols. According to \cref{eq:LR-prediction-rarepert}, the perturbation with the minimal $\Bar{\tau}_{P^*}$ is the preferred one. Finally, to obtain an efficient perturbation time interval, plug $\Psi_T(t\leq P^*)$ and $\Bar{\tau}_{P^*}$ into \cref{eq:LR-prediction-rarepert}, and evaluate the speedup for $\max(t_r,\Bar{\tau}_{P^*}) \leq P \leq P^*$. Choose the time interval that leads to the highest speedup. 
{In conclusion, if one plans to estimate the MFPT for a perturbation at a single rate, they might as well use our approach to predict whether lower rates would lead to higher accelerations at almost no added computational cost.}

\section{Summary and Conclusion}
We developed an approach based on a recent response theory by Keidar et al.~\cite{keidar2024universal} to design useful perturbations for accelerating the training of NNs.  
To that end, we treated the training as a first-passage process to a target test accuracy.
For the case of a CIFAR-10 classifier using the ResNet-18 architecture, we demonstrated that the unperturbed training test accuracy reached a QSS after a relaxation time of 20-30 epochs. We showed that, as a result, we can predict the mean FPT for a wide range of perturbation times from measurements at a single perturbation time. We focused on three perturbations: S\&P, SR and partial SR, but our method is general and can be used for other perturbations.
Lastly, we proposed a strategy to examine different kinds of perturbations and pick the optimal one and an efficient $P$. We showed that this strategy correctly selects S\&P as a better perturbation than partial SR or SR for a CIFAR-10 classifier and identified a useful perturbation time. 
 {We demonstrated the transferability of the methodology to other datasets, architectures, optimizers and even tasks (regression instead of classification).}
Our work provides a first-passage framework capable of identifying perturbations that result in a speedup when training an NN model up to a target accuracy.
It allows for a more rational design of perturbation protocols, based on physical insights. 

\section{Computational details}
The CIFAR-10 dataset contains 50,000 training and 10,000 test RGB images each belonging to one of 10 different classes. The number of images per class across the train and test sets is equal, and each image is 32$\times$32 pixels in size. For classification, we used a common modified version of the ResNet-18 architecture, because the original architecture was designed for much larger images. The modification includes reducing the kernel size of the first convolution layer from 7 to 3, the stride from 2 to 1, and the padding from 3 to 1. We also removed the subsequent max pooling layer to maintain everything else similar to the original architecture (as was done in the work of Zaidi et al.~\cite{zaidi2023does}). 

We did not use any data augmentation techniques except for standard data normalization (by the channel-wise mean and variance of the train images) for both train and test datasets. We initialized the weights and biases with the default initialization of Pytorch, i.e., uniform distribution, and used the FFCV library for faster training~\cite{leclerc2023ffcv}. We trained each model with an SGD optimizer with a learning rate of 0.02 {and batch size of 125}. Implementation of the perturbations was straightforward according to their definitions above. The test accuracy curves for each perturbative protocol with $P=20$ are presented in supporting information.

Raw data and example code to train the models and make the predictions are given in the associated repository: \url{https://github.com/Hirshberg-Lab/OptPerturbationsNN}

\section*{Acknowledgements}
B.H. acknowledges support from the Israel Science Foundation (grants No. 1037/22 and 1312/22), the Pazy Foundation of the IAEC-UPBC (grant No. 415-2023), and Tel Aviv University Center for Artificial Intelligence and Data Science (TAD). This project has received funding from the European Research Council (ERC) under the European Union’s Horizon 2020 research and innovation program (grant agreement No. 947731 to S.R.). S.M. acknowledges support from the Quantum Science and Technology Center of Tel Aviv University. B.H. and S.M. thank Sheheryar Zaidi for sharing his code. The authors thank Yohai Bar Sinai, Tomer Koren, and Rotem Widman for fruitful discussions.

\bibliography{refs}

\newpage
\renewcommand{\thefigure}{S\arabic{figure}}
\renewcommand{\thetable}{S\arabic{table}}
\renewcommand{\theequation}{S\arabic{equation}}
\setcounter{figure}{0} 
\setcounter{table}{0}
\setcounter{equation}{0}

\section*{Supporting Information: First-Passage Approach to Optimizing Perturbations for Improved Training of Machine Learning Models}

\subsection{Derivation of Eq.~3 of the main text}
This appendix shows the derivation of \cref{eq:LR-prediction}. Note that a similar derivation is presented in \cite{keidar2024universal}, but for exponentially distributed perturbation times instead of a constant $P$.
Using the law of total expectation on \cref{eq:T_P_cases_description}, the mean FPT is,
\begin{equation}
    \mathbb{E}[T_P] = \Pr(T \leq P)\mathbb{E}[T \mid T \leq P] + \Pr(T>P)\left(\mathbb{E}[P \mid T > P] + \mathbb{E}\left[\mathbb{E}[\tau_P(\boldsymbol{\theta})] \mid T > P\right]\right).
\end{equation}
We define
\begin{equation}
    \Bar{\tau}_P \equiv \mathbb{E}\left[\mathbb{E}[\tau_P(\boldsymbol{\theta})] \mid T > P\right] = 
    \int_{\Theta} \mathbb{E}[\tau_P(\boldsymbol{\theta})] \frac{G(\boldsymbol{\theta},P)}{\Pr(T>P)} \mathrm{d}\boldsymbol{\theta} = \int_{\Theta} \mathbb{E}[\tau_P(\boldsymbol{\theta})] \frac{G(\boldsymbol{\theta},P)}{\Psi_T(P)} \mathrm{d}\boldsymbol{\theta},
\end{equation}
as taking the expectation value of $\tau_P(\boldsymbol{\theta})$ first over all possible noise realizations of the stochastic training process after the perturbation was applied to a given state $\boldsymbol{\theta}$ at time $P$, and then over all possible $\boldsymbol{\theta}$ generated by the unperturbed process at time $P$ right before the perturbation has been applied. The mean FPT with the perturbation is then given by
\begin{equation}
    \mathbb{E}[T_P] = \Pr(T \leq P)\mathbb{E}[T \mid T \leq P] + \Pr(T>P)\left(\mathbb{E}[P \mid T > P] +\Bar{\tau}_P\right).
    \label{eq:SI-law_tot_expectation}
\end{equation}

Next, we write the survival function $\Psi_T(t)$ in terms of the probability mass function of $T$, $\mathcal{P}_T(t)$,
\begin{equation}
    \Psi_T(t) = \Pr(T>t) = 1 - \sum_{n=0}^t \mathcal{P}_T(n),
\end{equation}
such that $\mathcal{P}_T(t) = \Psi_T(t-1)-\Psi_T(t)$. With this definition, we can compute each conditional expectation in \cref{eq:SI-law_tot_expectation}. The first term is,
\begin{equation}
    \mathbb{E}[T \mid T \leq P]=\frac{1}{\Pr(T \leq P)}\sum_{t=0}^P t \mathcal{P}_T(t),
\end{equation}
where the normalization factor is $\Pr(T \leq P)$ due to the conditional average. Since the first term in the sum is zero, we get
\begin{equation}
    \mathbb{E}[T \mid T \leq P]=\frac{1}{\Pr(T \leq P)}\sum_{t=1}^P t \mathcal{P}_T(t)=\frac{1}{\Pr(T \leq P)}\sum_{t=1}^P t (\Psi_T(t-1)-\Psi_T(t)).
\end{equation}
By changing the summation index of the first sum, we can rewrite
\begin{equation}
    \mathbb{E}[T \mid T \leq P]=\frac{1}{\Pr(T \leq P)}\left( \sum_{t=0}^{P-1} (t+1)\Psi_T(t) -  \sum_{t=1}^P t\Psi_T(t)\right) = \frac{1}{\Pr(T \leq P)}\left( \sum_{t=0}^{P-1} \Psi_T(t)  - P\Psi_T(P) \right),
\end{equation}
which is the final expression for the first expectation value.
% \begin{equation}
% \begin{split}
%     \mathbb{E}[T \mid T \leq P]=\sum_{t=0}^{\infty} t \mathcal{P}_{T \mid T \leq P}(t) = \sum_{t=0}^P \frac{t \mathcal{P}_T(t)}{\Pr(T \leq P)} = \frac{1}{\Pr(T \leq P)}\sum_{t=1}^P t \mathcal{P}_T(t) = \\
%     =\frac{1}{\Pr(T \leq P)}\sum_{t=1}^P t (\Psi_T(t-1)-\Psi_T(t))  = \frac{1}{\Pr(T \leq P)}\left( \sum_{t=1}^P t \Psi_T(t-1) -  \sum_{t=1}^P t\Psi_T(t)\right) = \\
%     = \frac{1}{\Pr(T \leq P)}\left( \sum_{t=0}^{P-1} (t+1)\Psi_T(t) -  \sum_{t=1}^P t\Psi_T(t)\right) = \frac{1}{\Pr(T \leq P)}\left( \sum_{t=0}^{P-1} \Psi_T(t)  - P\Psi_T(P) \right) .
% \end{split}
% \end{equation}
The second term is,
\begin{equation}
    \mathbb{E}[P \mid T > P] =P= \frac{P\Psi_T(P)}{\Pr(T>P)}.
\end{equation}
% and the third term is the definition of $\Bar{\tau}_P$ according to \cref{eq:tau_P-definition},
% \begin{equation}
%     \mathbb{E}\left[\mathbb{E}\left[\tau_P(\boldsymbol{\theta})\right] \mid T > P\right] \equiv  \Bar{\tau}_P = \int_{\Theta} \mathbb{E}[\tau_P(\boldsymbol{\theta})] \frac{G(\boldsymbol{\theta},P)}{\Pr(T>P)} \mathrm{d}\boldsymbol{\theta} = \int_{\Theta} \mathbb{E}[\tau_P(\boldsymbol{\theta})] \frac{G(\boldsymbol{\theta},P)}{\Psi_T(P)} \mathrm{d}\boldsymbol{\theta}.
% \end{equation}
Plugging these two terms back into \cref{eq:SI-law_tot_expectation},
\begin{equation}
     \mathbb{E}[T_P] = \sum_{t=0}^{P-1} \Psi_T(t) - \cancel{P\Psi_T(P)} + \cancel{P\Psi_T(P)} +  \Psi_T(P)\Bar{\tau}_P ,
\end{equation}
results in \cref{eq:LR-prediction}.

\subsection{Computational details of testing the QSS hypothesis}
This appendix includes computational details and provides additional support for the QSS of the test accuracy during SGD training without perturbations. The CDF $F_t(A)$ of the accuracy $A$ at epoch $t$ was calculated according to,
\begin{equation}
    F_t(A)=\frac{1}{N(t)}\sum_{i=1}^{N(t)}I\{A_i(t)\leq A\}.
\end{equation}
Here, $N(t)$ is the number of models that did not reach the target accuracy at time $t$, $I\{E\}$ is the indicator function of the event $E$, and $A_i(t)$ is the accuracy of the $i$-th model at time $t$. The average CDF $\Bar{F}(A)$ of $A$ over epochs 20 to 100 was taken as an arithmetic average of the accuracy CDFs, i.e.,
\begin{equation}
    \Bar{F}(A)=\frac{1}{t_2-t_1+1}\sum_{t=t_1}^{t_2}F_t(A).
\end{equation}

\subsection{Other statistical tests}
\begin{figure}[H]
\begin{center}
\centerline{\includegraphics[width=0.9\linewidth]{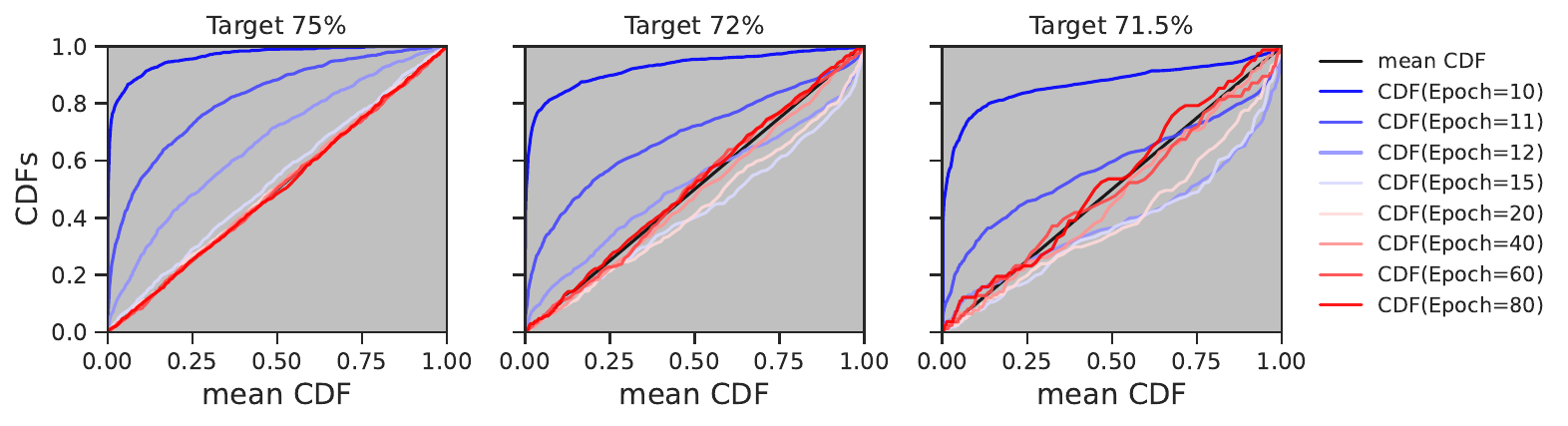}}
\caption{CDFs versus the mean CDF. Blue and red colors indicate CDFs before and after reaching a QSS respectively. }
\label{fig:SI_qss_distance}
\end{center}
\end{figure}

In \cref{fig:SI_qss_distance} we plot the CDFs of the test accuracy versus the values of the mean CDF of $A$ over epochs 20 to 100. The mean CDF is plotted in black for qualitative comparison between other CDFs (see legend). 

\begin{figure}[H]
\begin{center}
\centerline{\includegraphics[width=0.5\linewidth]{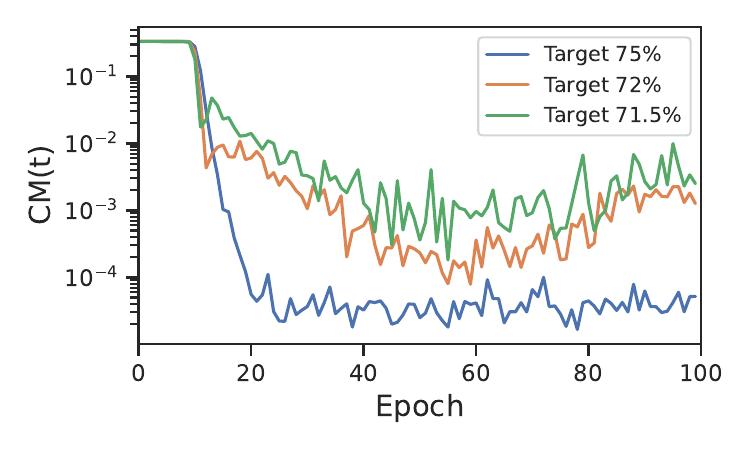}}
\caption{Cram{\'e}r–von Mises criterion}
\label{fig:SI_qss_mc_test}
\end{center}
\end{figure}

In \cref{fig:SI_qss_mc_test} we compare the CDFs of the test accuracy $A$ to the average CDF of $A$ over epochs 20 to 100 with another measure, the Cram{\'e}r–von Mises (CM) criterion. The CM criterion at epoch $t$ is calculated as follows,
\begin{equation}
    CM(t) = \int_{0}^{1} \left( F_t(A) - \Bar{F}(A) \right)^2\mathrm{d}\Bar{F}(A).
\end{equation}
Here, $\Bar{F}(A)$ is the average CDF of $A$ over epochs 20 to 100.
In practice, we take the square of the difference between each curve of \cref{fig:SI_qss_distance} and the mean CDF, and integrate along the x-axis.

\subsection{Training with and without perturbations}
\begin{figure}[H]
\begin{center}
\centerline{\includegraphics[width=0.55\linewidth]{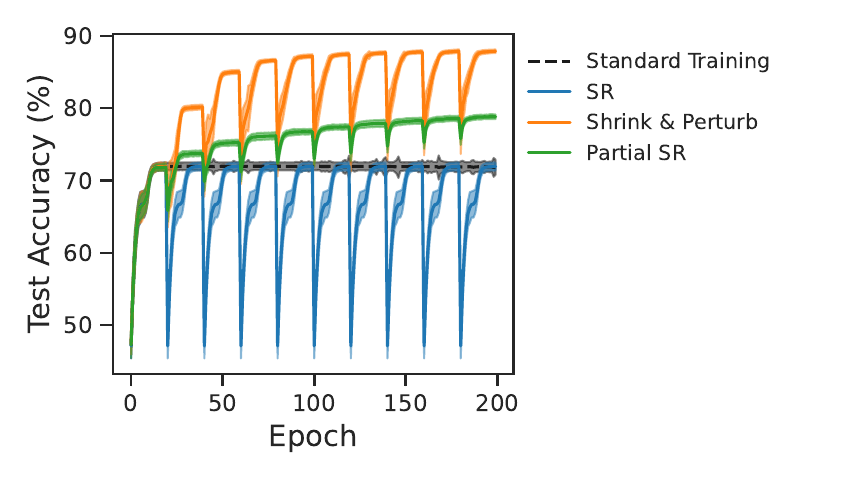}}
\caption{Test accuracy of different training protocols with perturbation time $P=20$ epochs. Standard training in black, SR in blue, S\&P in orange, and partial SR in green. }
\label{fig:test_acc}
\end{center}
\end{figure}

This appendix shows in \cref{fig:test_acc} the mean test accuracy curves of standard SGD training (black), and training with SR (blue), S\&P (orange), and partial SR (green).

\subsection{Transferability across datasets, optimizers, architectures and tasks}
{In this section, we demonstrate that the QSS is not a unique feature of the CIFAR-10 and ResNet-18 dataset and architecture. We test additional datasets, architecture, optimizer, and even task (classification versus regression). The following subsections show three experimental tests: 1) CIFAR-100 classification with all other hyperparameters and setting as in the main text, 2) CIFAR-100 classification, but using SGD with momentum, and 3) a regression phase retrieval task on the MNIST dataset~\cite{deng2012mnist} using a fully connected achitechture and the Adam optimizer~\cite{kingma2014adam}.}

\subsubsection{CIFAR-100 Classification}
\begin{figure}[H]
\begin{center}
\centerline{\includegraphics[width=0.55\linewidth]{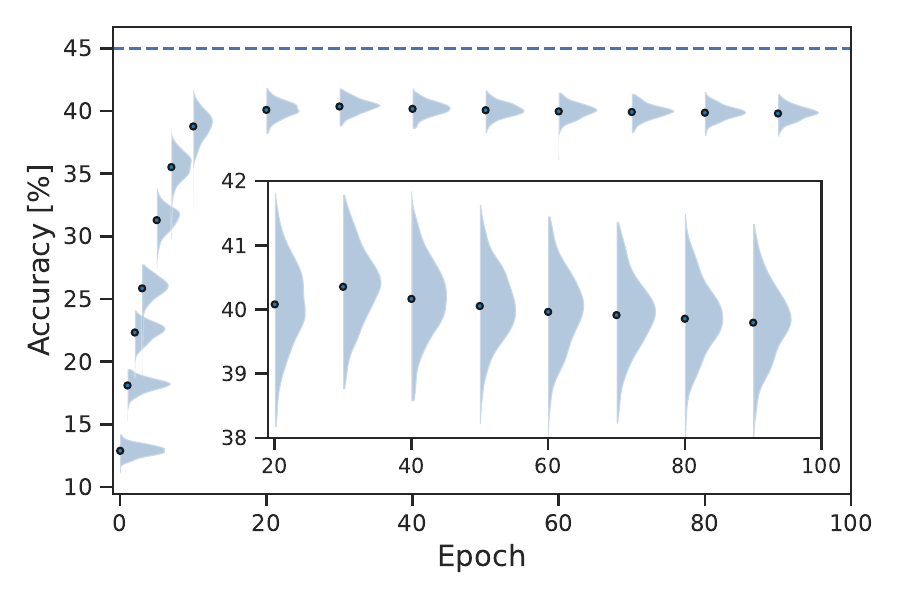}}
\caption{{Experimental evidence for a quasi-steady-state on the CIFAR-100 dataset. In blue are violin plots of the density distributions of the test accuracy as a function of the number of epochs, for models that did not reach the target. Dots represent the mean accuracy and the areas are the distributions rotated by $90^{\circ}$. The dashed blue line is the target accuracy set to 45\%.}}
\label{fig:acc_dist_cifar100_No_Momentum}
\end{center}
\end{figure}

{
In~\cref{fig:acc_dist_cifar100_No_Momentum}, we show that the training process on the CIFAR-100 dataset also reaches a QSS. We used the same optimizer, architecture, and learning rate as was used for the CIFAR-10 dataset.
}

\begin{figure}[H]
\begin{center}
\centerline{\includegraphics[width=0.4\linewidth]{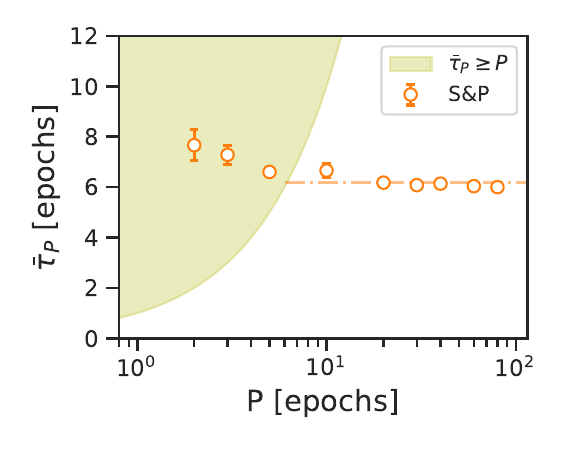}}
\caption{{The mean residual time $\Bar{\tau}_P$ to reach 45\% test accuracy versus the perturbation time, for CIFAR-100 classification task. The values of $\Bar{\tau}_P$ for S\&P (orange circles) are approximately constant when $P>\Bar{\tau}_P$ (outside the yellow area). Dashed dotted line is the average over all $\Bar{\tau}_P<P$.}}
\label{fig:SI_tau_P_vs_P_cifar100_No_Momenrum}
\end{center}
\end{figure}

{Since the training process reaches a QSS we predict that $\Bar{\tau}_P$ is independent of $P$ for a wide range according to~\cref{eq:LR-prediction-rarepert}. To verify it, we applied the S\&P perturbation to the training trajectories that did not reach the target prior to time $P$ (see~\cref{fig:SI_tau_P_vs_P_cifar100_No_Momenrum}).}

\begin{figure}[H]
\begin{center}
\centerline{\includegraphics[width=0.4\linewidth]{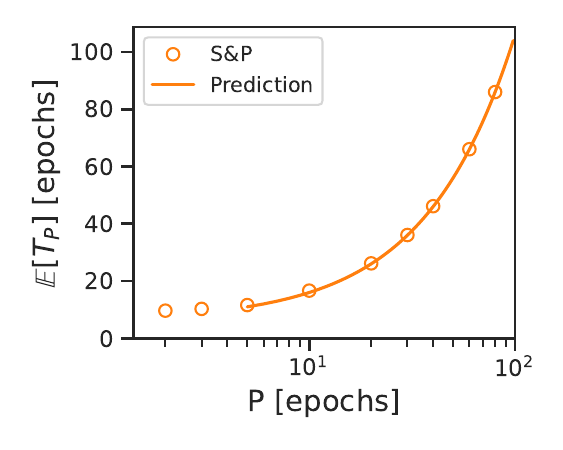}}
\caption{{The perturbed mean FPT to reach 45\% test accuracy on the CIFAR-100 classification. The orange line is the theoretical predictions of \cref{eq:LR-prediction-rarepert} for S\&P. For the predictions, we used the value of $\Bar{\tau}_{P^*}$ for $P^*=80$.}}
\label{fig:SI_Tp_cifar100_No_Momentum}
\end{center}
\end{figure}

{Finally, we predict the mean perturbed FPT $\mathbb{E}[T_P]$ to reach a target test accuracy of 45\%, shown in ~\cref{fig:SI_Tp_cifar100_No_Momentum}.}

\subsubsection{CIFAR-100 classification - training with momentum}

\begin{figure}[H]
\begin{center}
\centerline{\includegraphics[width=0.55\linewidth]{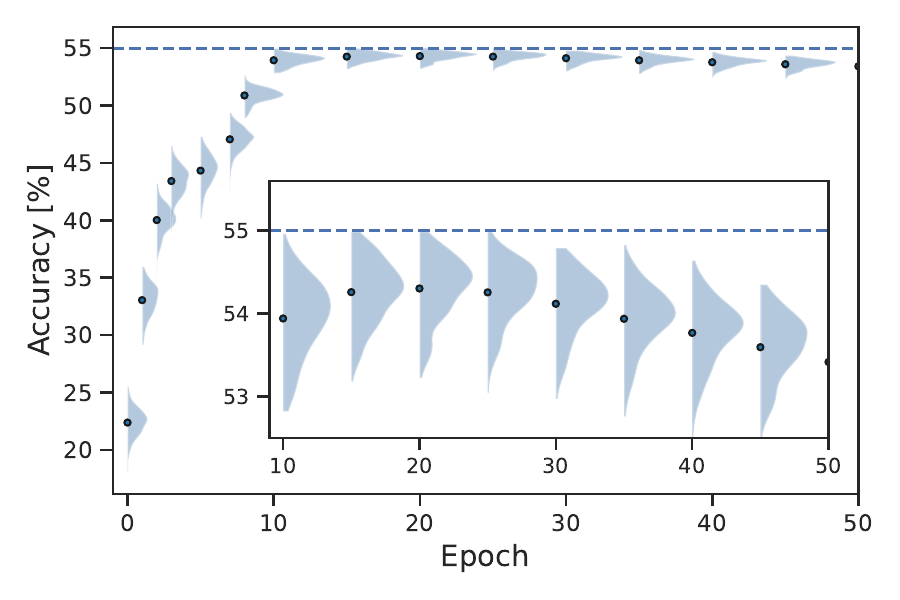}}
\caption{{Experimental evidence for a quasi-steady-state on the CIFAR-100 dataset with momentum. In blue are violin plots of the density distributions of the test accuracy as a function of the number of epochs, for models that did not reach the target. Dots represent the mean accuracy and the areas are the distributions rotated by $90^{\circ}$. The dashed blue line is the target accuracy set to 55\%.}}
\label{fig:SI_acc_dist_cifar100}
\end{center}
\end{figure}

{
In~\cref{fig:SI_acc_dist_cifar100}, we show that the training process with a momentum term, on the CIFAR-100 dataset reaches a QSS. We used the same architecture and learning rate as was used for the CIFAR-10 dataset. We used a momentum value of 0.9.
}

\begin{figure}[H]
\begin{center}
\centerline{\includegraphics[width=0.4\linewidth]{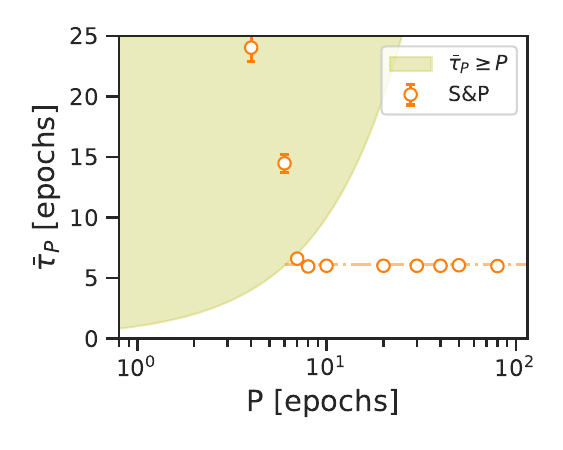}}
\caption{{The mean residual time $\Bar{\tau}_P$ to reach 55\% test accuracy versus the perturbation time for the CIFAR-100 classification task when trained with momentum. The values of $\Bar{\tau}_P$ for S\&P (orange circles) are approximately constant when $P>\Bar{\tau}_P$ (outside the yellow area). Dashed dotted line is the average over all $\Bar{\tau}_P<P$.}}
\label{fig:SI_tau_P_vs_P_cifar100}
\end{center}
\end{figure}

{Similarly to the case without momentum we show that $\Bar{\tau}_P$ is independent of $P$ for a wide range in~\cref{fig:SI_tau_P_vs_P_cifar100}.}

\begin{figure}[H]
\begin{center}
\centerline{\includegraphics[width=0.4\linewidth]{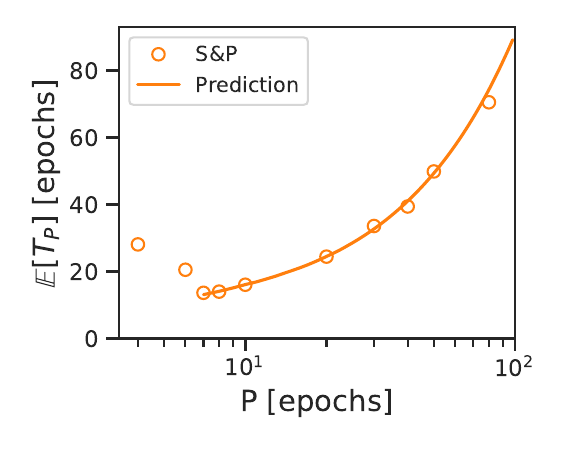}}
\caption{{The perturbed mean FPT to reach 55\% test accuracy on the CIFAR-100 classification with momentum. The orange line is the theoretical predictions of \cref{eq:LR-prediction-rarepert} for S\&P. For the predictions, we used the value of $\Bar{\tau}_{P^*}$ for $P^*=50$.}}
\label{fig:SI_Tp_cifar100}
\end{center}
\end{figure}

{Finally, we predict the mean perturbed FPT $\mathbb{E}[T_P]$ to reach a target test accuracy of 55\%, shown in~\cref{fig:SI_Tp_cifar100}.}

\subsubsection{Regression task - phase retrieval of images}

\begin{figure}[H]
\begin{center}
\centerline{\includegraphics[width=0.9\linewidth]{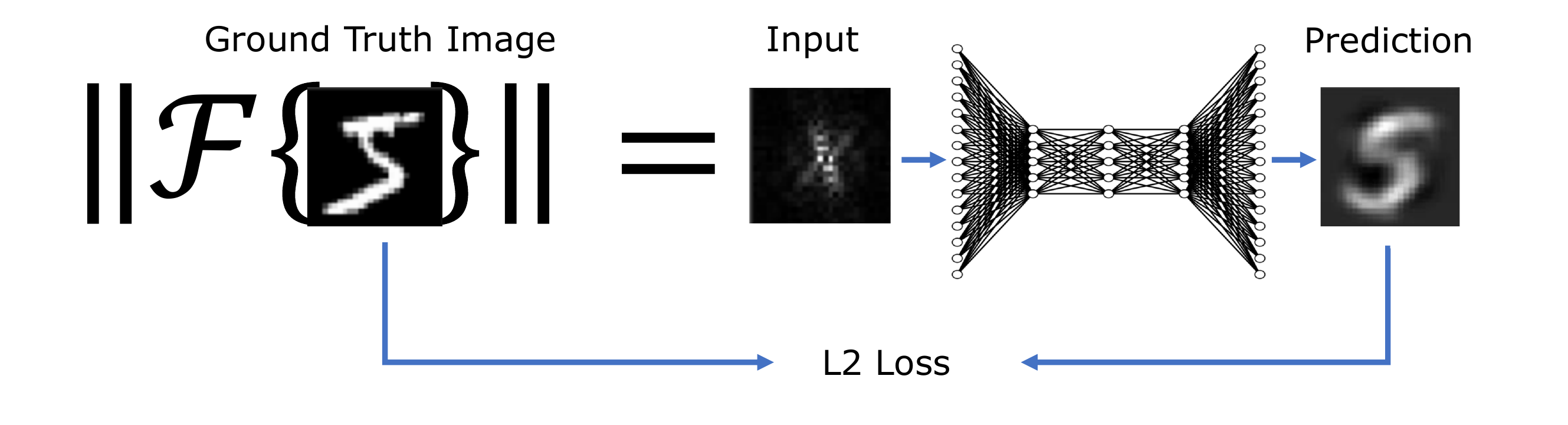}}
\caption{{Example of the phase retrieval problem settings. We take the absolute value over the Fourier transform of the MNIST images, to create the inputs for the model. The model learns to reconstruct images from their magnitude by minimizing the L2 loss between the ground truth and the output images. We use an L2 loss that is invariant to 180\degree rotation.}}
\label{fig:SI_PR_example}
\end{center}
\end{figure}

{To demonstrate a QSS for a regression task, we trained a fully-connected NN to reconstruct images from their Fourier-transformed magnitude, i.e., the absolute value of their Fourier transform. We used a similar problem setting as in Dekel et al.~\cite{dekel2022pr}. The input of the network is created by taking the absolute value over the Fourier transform of the MNIST images (see \cref{fig:SI_PR_example}). The model learns to reconstruct the corresponding image by minimizing an L2 loss between its output and the ground truth image.}

{To elevate the difficulty of this problem we swapped the roles between the test and train datasets of MNIST. We used the Adam optimizer with a learning rate of 0.1 and batch size of 1000. The architecture of our model contains four fully connected layers (784$\times$256$\times$256$\times$784). Between every two layers, there is a batch normalization layer followed by a parametric ReLU activation. Also, we used an L2 loss that is invariant to 180\degree \, rotation of the predicted image. This is because both images would have the same magnitude.}

\begin{figure}[H]
\begin{center}
\centerline{\includegraphics[width=0.85\linewidth]{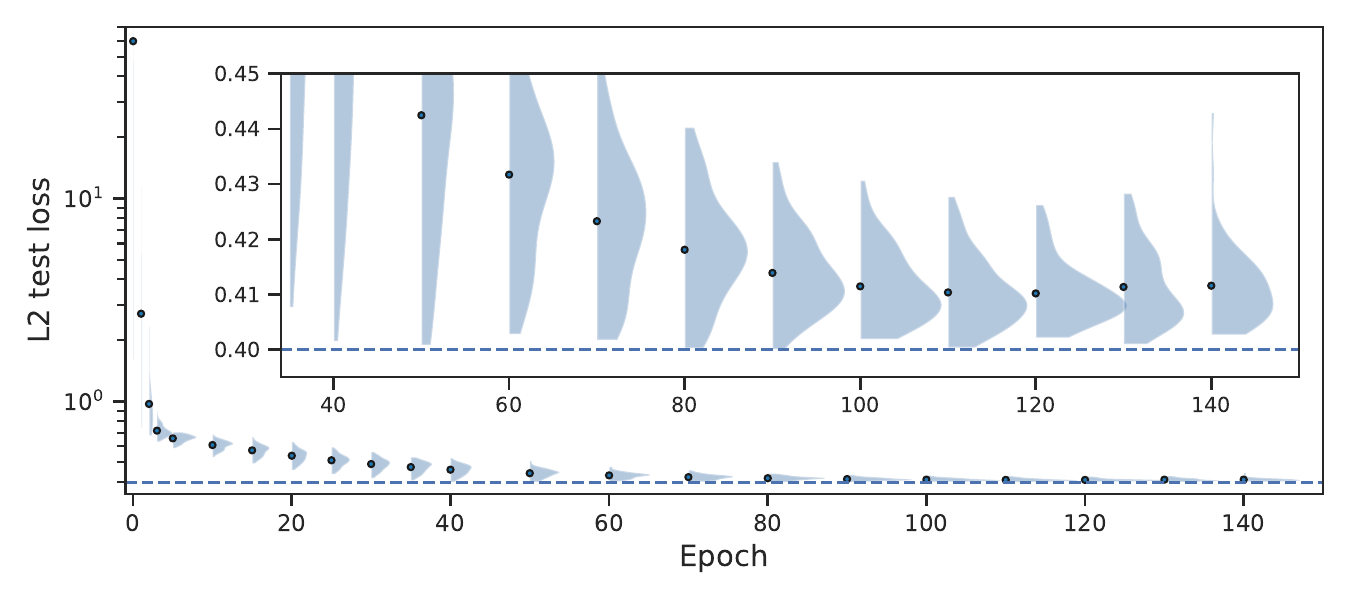}}
\caption{{Experimental evidence for a quasi-steady-state for a regression task. The target was set on 0.4 L2 test loss.}}
\label{fig:SI_regression_QSS}
\end{center}
\end{figure}

{To demonstrate a QSS in the learning process of this regression task, we observe the distribution of the L2 test loss instead of ``test accuracy'' that was used for the classification tasks (see~\cref{fig:SI_regression_QSS}.}

\begin{figure}[H]
\begin{center}
\centerline{\includegraphics[width=0.4\linewidth]{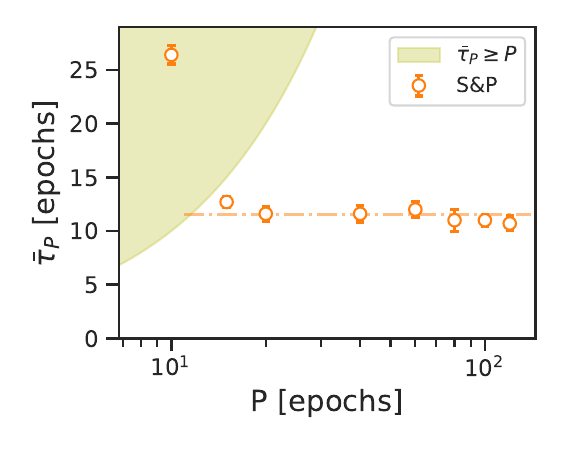}}
\caption{{The mean residual time $\Bar{\tau}_P$ versus the perturbation time for the regression task. The values of $\Bar{\tau}_P$ for S\&P (orange circles) are approximately constant when $P>\Bar{\tau}_P$ (outside the yellow area). Dashed dotted line is the average over all $\Bar{\tau}_P<P$.}}
\label{fig:SI_tau_P_vs_P_mnistPR}
\end{center}
\end{figure}

{As in the classification case, we demonstrate that $\Bar{\tau}_P$ is independent of $P$ for a wide range in~\cref{fig:SI_tau_P_vs_P_mnistPR}. We used the S\&P protocol but with different hyperparameters: $\lambda$=0.8 for shrinking and $\gamma=$0.05 for perturbing.}

\begin{figure}[H]
\begin{center}
\centerline{\includegraphics[width=0.4\linewidth]{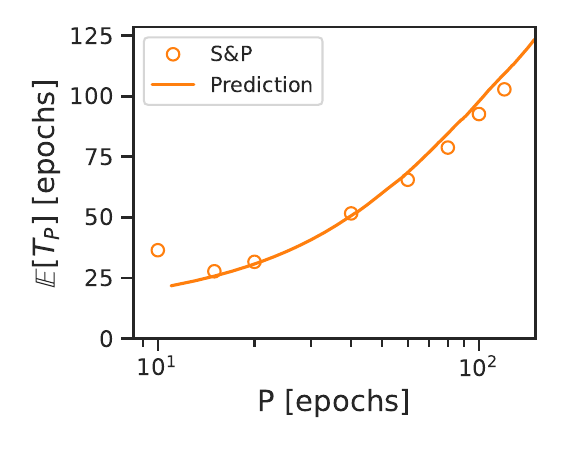}}
\caption{{The perturbed mean FPT to reach 0.4 L2 test loss with the Adam optimizer for the regression task. The orange line is the theoretical prediction of \cref{eq:LR-prediction-rarepert} for S\&P. For the predictions, we used the value of $\Bar{\tau}_{P^*}$ for $P^*=120$.}}
\label{fig:SI_Tp_mnistPR}
\end{center}
\end{figure}

{Finally, we predict the mean perturbed FPT $\mathbb{E}[T_P]$ to reach a target loss value of 0.4, shown in~\cref{fig:SI_Tp_mnistPR}.}

\end{document}